\documentclass{article}

\usepackage{microtype}
\usepackage{graphicx}
\usepackage{subfigure}
\usepackage{booktabs} 

\usepackage{hyperref}
\usepackage{paralist}

\usepackage[accepted]{icml2023}

\usepackage{amsmath}
\usepackage{amssymb}
\usepackage{mathtools}
\usepackage{amsthm}

\usepackage[capitalize,noabbrev]{cleveref}

\theoremstyle{plain}

\theoremstyle{definition}

\theoremstyle{remark}

\usepackage[textsize=tiny]{todonotes}

\usepackage{url}
\usepackage{multirow}
\usepackage{graphicx}
\usepackage{booktabs}
\usepackage{threeparttable}
\usepackage{bbm}
\usepackage{wrapfig}
\usepackage{caption}

\icmltitlerunning{Scaling Up Dataset Distillation to ImageNet-1K with Constant Memory}

\begin{document}

\twocolumn[
\icmltitle{Scaling Up Dataset Distillation to ImageNet-1K with Constant Memory}



\icmlsetsymbol{equal}{*}

\begin{icmlauthorlist}
\icmlauthor{Justin Cui}{UCLA}
\icmlauthor{Ruochen Wang}{UCLA}
\icmlauthor{Si Si}{Google Research}
\icmlauthor{Cho-Jui Hsieh}{UCLA}
\end{icmlauthorlist}

\icmlaffiliation{UCLA}{Department of Computer Science, University of California, Los Angeles}
\icmlaffiliation{Google Research}{Google Research}

\icmlcorrespondingauthor{Justin Cui}{justincui@ucla.edu}
\icmlcorrespondingauthor{Cho-Jui Hsieh}{chohsieh@cs.ucla.edu}

\icmlkeywords{Machine Learning, Dataset Distillation, Dataset Condensation, TESLA, ICML}

\vskip 0.3in
]



\printAffiliationsAndNotice{} 

\begin{abstract}
Dataset Distillation is a newly emerging area that aims to distill large datasets into much smaller and highly informative synthetic ones to accelerate training and reduce storage. 
Among various dataset distillation methods, trajectory-matching-based methods (MTT) have achieved SOTA performance in many tasks, e.g., on CIFAR-10/100. However, due to exorbitant memory consumption when unrolling optimization through SGD steps, MTT fails to scale to large-scale datasets such as ImageNet-1K. 
Can we scale this SOTA method to ImageNet-1K and does its effectiveness on CIFAR transfer to ImageNet-1K? 
To answer these questions, we first propose a procedure to exactly compute the unrolled gradient with constant memory complexity, which allows us to scale MTT to ImageNet-1K seamlessly with $\sim 6$x reduction in memory footprint.
We further discover that it is challenging for MTT to 
handle datasets with a large number of classes, and propose a novel soft label assignment that drastically improves its convergence. The resulting algorithm sets new SOTA on ImageNet-1K: we can scale up to 50 IPCs (Image Per Class) on ImageNet-1K on a single GPU (all previous methods can only scale to 2 IPCs on ImageNet-1K), leading to the best accuracy (only 5.9\% accuracy drop against full dataset training) while utilizing only 4.2\% of the number of data points - an 18.2\% absolute gain over prior SOTA. Our code is available at~\href{https://github.com/justincui03/tesla}{https://github.com/justincui03/tesla}
\end{abstract}

\vspace{-20pt}
\section{Introduction}
\label{submission}
In this paper, we study the problem of dataset distillation, where the goal is to distill a large dataset into a small set of synthetic samples, such that models trained with the synthetic samples can achieve competitive performance compared with training on the whole dataset~\cite{wang2018dataset}. Different from core-set selection~\cite{wolf2011facility, rebuffi2017icarl, castro2018end}, synthetic samples are learned freely in the continuous space instead of being selected from the original dataset, so they often achieve better performance in the regime with higher compression rates.
Due to the importance of compressing a large dataset into smaller ones, many algorithms have been proposed in the past few years, including Gradient Matching~\cite{zhaodsa}, Distribution Matching~\cite{zhaodm}, KIP~\cite{nguyen2021dataset} and Matching Training Trajectories (MTT)~\cite{cazenavette2022dataset}. 
Recently, a line of methods based on matching trajectory has demonstrated state-of-the-art performance on smaller datasets~\cite{dcbench, cazenavette2022dataset}: When using only 50 synthetic images per class which yields a 100x compression rate, 
MTT only incurs 13.4\% accuracy drop compared to training on the whole CIFAR-10 dataset.

Despite achieving state-of-the-art performance, MTT cannot scale to large datasets due to its huge GPU memory requirement~\cite{zhou2022dataset,cazenavette2022dataset,dcbench}. This is fundamentally due to the objective function of MTT, which unrolls $T$ SGD updates with synthetic images and matches the resulting weights with a reference point (obtained by training on the original dataset). Since this objective function unrolls $T$ optimization steps, back-propagating requires expanding and storing $T$ gradient computational graphs in GPU memory and is prohibitive in large-scale problems. For instance, unrolling $T=30$ steps on CIFAR-10 requires 47GB GPU memory~\cite{cazenavette2022dataset} with IPC 50, and thus it runs out of memory when scaling to ImageNet-1K. This has become the main issue when scaling MTT to large problems.

In this paper, we propose a memory-efficient version of MTT, which only requires storing a single gradient computational graph even when unrolling $T$ steps. This reduces the GPU memory complexity of MTT with respect to number of unrolled steps from linear to constant, while achieving identical performances and with only marginal computational overhead. This is done by a novel procedure to cache and rearrange the gradient computation of the trajectory matching loss. Equipped with the proposed method, we are able to scale MTT to ImageNet-1K with 1, 2, 10, 50 IPCs. In the literature, there exists only one most recent paper that scales to ImageNet-1K with IPC up to 2, but it encounters memory and runtime issues (\cref{sec.complexity}) when scaling to larger IPCs~\cite{zhou2022dataset}.

When applying memory-efficient MTT to ImageNet-1K, we observe extremely slow convergence with sub-optimal performance when assigning hard labels to the synthetic images.
We hypothesize that the missing ingredient is to use soft labels for synthesizing samples when dealing with a large number of classes, as soft labels allow information sharing across different classes.
This is also observed in FrePo~\cite{zhou2022dataset}, which jointly optimizes labels and synthetic images.
However, allowing labels to be freely learned also makes the inner optimization of our matching-based method harder to solve, resulting in only marginal performance gains.
To overcome this issue, we propose a soft label assignment (SLA) method that directly leverages the existing set of reference points (teacher models) in MTT for label assignment.
Concretely, at every iteration, we pass the synthetic images to the sampled teacher model, and directly use its generated soft labels to guide the training of synthetic images.
The proposed SLA is train-free and introduces zero hyperparameters. Empirically, the resulting algorithm significantly outperforms the original MTT on ImageNet-1K. Our contributions can be summarized below: 
\begin{compactitem}
\item We propose a novel method to reduce the memory usage of MTT from $\mathcal{O}(T)$ to $\mathcal{O}(1)$, where $T$ is the matching steps in MTT. This allows MTT to seamlessly scale to large datasets such as ImageNet-1K. 
\item We found assigning soft labels to synthetic images is crucial when scaling to datasets with a large number of labels (e.g., ImageNet-1K). However, naively learning soft labels works poorly for MTT. To overcome this issue, we propose Soft Label Assignment (SLA) - a novel hyperparameter-free method that directly injects soft labels into MTT from its reference models.
\item By combining the above-mentioned innovations, our method, codenamed TESLA (TrajEctory matching with Soft Label Assignment), outperforms state-of-the-art results under 1 and 2 IPCs on ImageNet-1K. Further, TESLA is the first in the field that scales to ImageNet-1K with IPC=10 and 50, 25X times larger than the next competitor.
\end{compactitem}

\section{Related Work}
The dataset distillation problem was first formally proposed by \cite{wang2018dataset}, where the goal is to compress a large dataset into a small set of synthetic samples. Although the compression stage could be computationally intensive, the distilled synthetic set can be used in multiple applications such as continuous learning~\cite{wang2018dataset,zhaodc}, federated learning~\cite{zhou2020distilled,xiong2022FedDM} and neural architecture search~\cite{zhaodsa, wang2021rethinking}. Many data distillation algorithms have been proposed in the past few years, and they can be roughly categorized into two types: matching-based approaches and kernel-based approaches. 

\paragraph{Matching-based Approaches:}
\cite{zhaodc,zhaodsa} propose to generate synthetic datasets by matching gradients between two surrogate models trained on distilled dataset and the real dataset.  
However, matching gradient requires high memory usage and computation time, so 
\cite{zhaodm} further proposes to match the features generated by the surrogate model.
Other recent works \cite{pmlr-v162-kim22c,deng2022remember,lee2022dataset} focus on learning lower-resolution synthetic images and upsampling, which can be applied to most of the existing methods and thus are orthogonal to our work.

Recently, \cite{cazenavette2022dataset} proposed a data distillation method based 
 on Matching Training Trajectories. This method achieves state-of-the-art performance on all the medium-sized datasets (e.g., CIFAR-10, CIFAR-100) and furthermore, according to DC-BENCH~\cite{dcbench}, MTT outperforms other works on not only accuracy but also transferability and stability (under various kinds of augmentations and IPC settings). 
The idea of MTT and its scalability issues will be discussed in the next section. 

\noindent\textbf{Kernel-based Approaches:}
Dataset distillation is intrinsically a bi-level optimization problem, where the inner optmization computes the model parameters given the synthetic dataset, and the outer optimization optimizes the synthetic dataset to minimize the loss of the resulting model. 
Inspired by the Neural Tangent Kernel (NTK), \cite{nguyen2020dataset, nguyen2021dataset} 
 use kernel ridge regression with NTK to obtain a closed form solution for the inner problem, and thus reducing the original bi-level optimization into a single-level optimization problem. This method, known as KIP, requires thousands of GPU hours due to the NTK computation.
To reduce the computational cost, FrePo~\cite{zhou2022dataset} only considers the neural network parameters of the last layer as learnable while keeping other parameters fixed. With this approximation, FrePo is able to obtain a closed form solution of ridge regression. Although FrePo is faster than KIP, it still requires storing all computational graphs and a heavy matrix inversion operation. Therefore it has difficulty scaling to larger IPCs.

\section{Background}
\subsection{Matching Training Trajectories}
Matching Training Trajectories (MTT) \cite{cazenavette2022dataset} proposes to generate the synthetic dataset by directly matching the model parameters trained using synthetic datasets with those trained on real datasets, which leads to the following loss function: 
\begin{equation}
    \label{eq:tm_loss}
    \mathcal{L} = \| \hat{\theta}_{t+T} - \theta_{t+M}^* \|^2_2 / \| \theta_t^{*} - \theta_{t+M}^* \|_2^2.
\end{equation}
Here $\theta_t^*$ represents the model parameter trained on real images at step $t$. Starting from $\theta_t^*$, 
 $\hat{\theta}_{t+T}$ denotes the model parameter trained on the synthetic dataset after $T$ steps and $\theta^*_{t+M}$ denotes the model parameter trained on the real dataset after $M$ steps. The goal of MTT is to have models trained on synthetic dataset with $T$ steps match the same results with teacher models trained from much more $M$ steps on real data and usually $T\ll M$.
We assume the model is updated by the standard SGD rule as below, where $\beta$ is the student model learning rate: 
\begin{equation}
\label{eq:tm_update}
\hat{\theta}_{t+i+1} = \hat{\theta}_{t+i} - \beta\nabla\ell(\hat{\theta}_{t+i}; \tilde{X}_{i}). 
\end{equation}
Here $\tilde{X}_{i}$ is a batch of (potentially augmented) synthetic images sampled from synthetic dataset $\tilde{X}$(See Appendix \cref{fig:mtt} for an illustration). 

\subsection{Scability of the current MTT method}
Although MTT achieves state-of-the-art performances on small datasets, it fails to scale to real-world large datasets such as ImageNet-1K similar to most existing condensation methods~\cite{zhaodsa, zhaodm, nguyen2020dataset, nguyen2021dataset, wang2022cafe}. The poor scalability significantly limits its practicality.

Before presenting our method, we start by demonstrating that the bottleneck of MTT's scalability lies in its unrolled gradient computation. To show this, we expand the MTT loss function defined in~\cref{eq:tm_loss} as follows. $\theta_t^{*}$ and $\theta_{t+M}^{*}$ in the denominator are all from pretrained model trajectories, thus they can be treated as constants. Unrolling $T$ steps of SGD update leads to 
\begin{align}
    \hat{\theta}_{t+T} &= \theta_{t}^* - \beta\nabla_{\theta} \ell(\theta_{t}^{*};\tilde{X}_0) - \beta\nabla_{\theta} \ell(\hat{\theta}_{t+1};\tilde{X}_{1}) - ... \nonumber\\
    &- \beta\nabla_{\theta} \ell(\hat{\theta}_{t+T-1};\tilde{X}_{T-1}). \label{eq:expansion}
\end{align}
Plugging this back into~\cref{eq:tm_loss}, it becomes
\begin{align}
    \label{eq:square_loss_ref}
    &\|\hat{\theta}_{t+T}-\theta_{t+M}^*\|^2_2=
    \nonumber\\
    &\hspace{1cm}\|\theta_{t}^* - \beta\overset{T-1}{\underset{i=0}{\sum}}\nabla_{\theta} \ell(\hat{\theta}_{t+i};\tilde{X}_i) - \theta_{t+M}^*\|^2_2.
\end{align}

To minimize $\mathcal{L}$, MTT needs to take the derivative of~\cref{eq:square_loss_ref} w.r.t. synthetic images. This involves computing and storing the computation graphs for $T$ high order gradient terms, where $T$ is the length of the trajectory. As the dataset size increases, the number of steps to train a model (trajectory length) also increases linearly, assuming everything else stays the same. As a result, \textbf{the GPU memory requirement for optimizing MTT loss becomes extremely large on larger datasets.} Also naively reducing/fixing matching step length leads to suboptimal performance, as redundant information can be encoded into multiple images~\cite{cazenavette2022dataset}.

\section{ Our proposed method}
In this section, we will discuss how to handle the scalablity issue of MTT, and propose our method: TrajEctory matching
with Soft Label Assignment (TESLA).

\subsection{MTT with constant memory}
\label{sec.constant_memory}
In this subsection, we present a computationally efficient way to resolve the scalability issue of MTT while obtaining the same performances.
Surprisingly, we found that with a careful rearrangement of the computation, the memory complexity of MTT can be reduced from linear to constant w.r.t. the 
number of trajectory matching steps - storing only one computational graph. We have also empirically verified that the proposed approach leads to identical performances as the original unrolled version of MTT. 

As we are computing the squared error of student and teacher model, \cref{eq:square_loss_ref} can be further expanded as
\begin{align}
\|\hat{\theta}_{t+T} - \theta_{t+M}^*\|_2^2 \ = \ & 
 2\beta(  \theta_{t+M}^* - \theta_t^*)^T(\overset{T-1}{\underset{i=0}{\sum}}\nabla_{\theta} \ell(\hat{\theta}_{t+i};\tilde{X}_i))
  \nonumber \\  +\beta^2 &\|\sum_{i=0}^{T-1} \nabla_\theta \ell(\hat{\theta}_{t+i}; \tilde{X}_i) \|^2 + C, \label{eq:111}
\end{align}
 where $C=\|\theta_{t}^* - \theta_{t+M}^*\|_2^2$ is a constant so can be ignored for gradient computation. It can be noticed that each term in the first  summation only involves the gradient of a single batch, so their gradients can be calculated sequentially without maintaining $N$ computational graphs. Only the second term 
 $\|\sum_{i=0}^{T-1} \nabla_\theta \ell(\hat{\theta}_{t+i}; \tilde{X}_i) \|^2$ involves information from multiple batches together. 
 
\begin{figure*}
    \centering
        \includegraphics[width=0.9\textwidth]{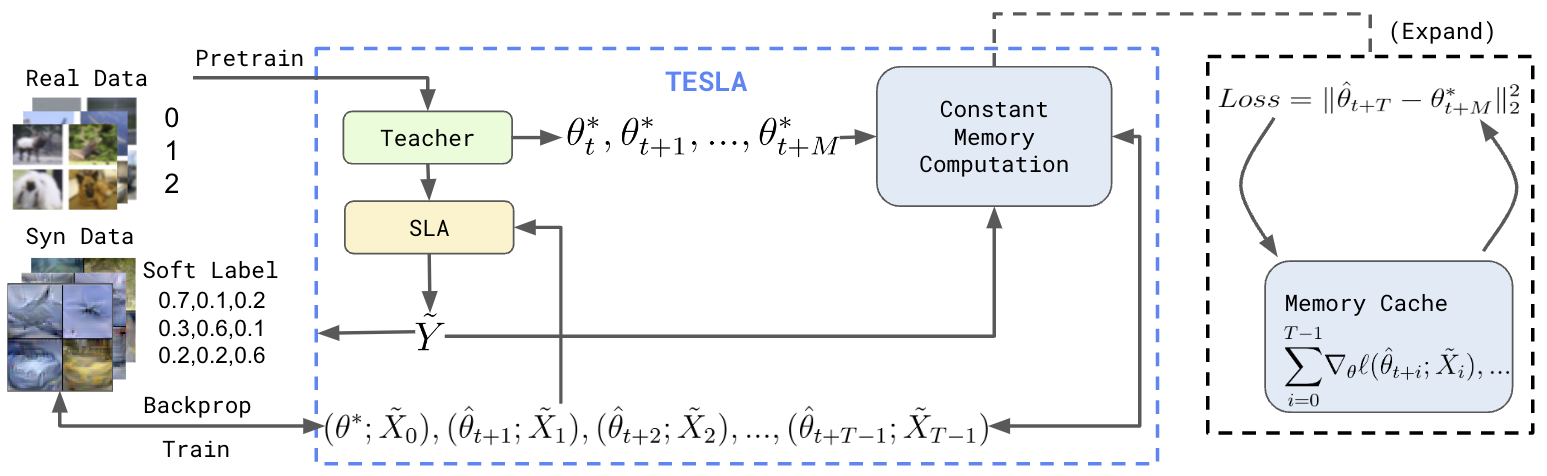}
        \caption{Illustration of trajectory matching with soft label assignment (TESLA). Our method differs from MTT in two aspects: 1). The constant memory computation module calculates the exact trajectory matching loss as MTT but with constant memory complexity. 2). The train-free Soft-Label Assignment (SLA) module leverages the pretrained teacher to distill soft labels to the synthetic data points.}
        \label{fig:tesla}
        \vspace{-10pt}
\end{figure*}

\paragraph{Backpropagating from MTT loss to a mini-batch}
Computing the gradient with respect to $\{\tilde{X}_i\}_{i=0}^{T-1}$ leads to 
\begin{align}
\frac{\partial \|\hat{\theta}_{t+T}-\theta^*_{t+M}\|_2^2}{\partial \tilde{X}_i} =& 
2\beta(\theta_{t+M}^* - \theta_t^*)^T \frac{\partial}{\partial \tilde{X}_i}\nabla_{\theta} \ell(\hat{\theta}_{t+i};\tilde{X}_i) \nonumber\\
&+ 2\beta^2 G^T \frac{\partial}{\partial \tilde{X}_i} \nabla_{\theta} \ell(\hat{\theta}_{t+i};\tilde{X}_{i}), \label{eq:gradient} 
\end{align}
where $G=\sum_{i=0}^{T-1} \nabla_{\theta} \ell(\hat{\theta}_{t+i};\tilde{X}_{i})$.
Our key finding of this expansion is as follows: since G can be precomputed in MTT (1st for-loop in Algorithm \ref{alg:algorithm}) while updating $\hat{\theta}$, the only computation graph required for computing \eqref{eq:gradient} is the gradient on batch $\tilde{X}_i$: $\nabla_{\theta} \ell(\hat{\theta}_{t+i};\tilde{X}_{i})$.
As the computation graph of $ \nabla_{\theta} \ell(\hat{\theta}_{t+i};\tilde{X}_{i})$ is not required in the derivative of any other batch $\tilde{X}_{j \neq i}$, it can be released right after use.
Therefore, the memory complexity of computing MTT loss becomes independent of the step $t$, thus constant w.r.t. $T$.

\paragraph{Backpropagating from a mini-batch to synthetic images}
To further backpropagate from a mini-batch to the synthetic images, we need to define the connections between batches $\{\tilde{X}_i\}_{i=0}^{T-1}$ and synthetic images $\{x_j\}_{j=1}^n$.
Assume each batch $i$ contains $B$ samples denoted as  $\tilde{X}_i = [\tilde{X}_{i,1}, \dots, \tilde{X}_{i, B}]$ (concatenation of $B$ vectors) and for all $b=1, \dots, B$, $\tilde{X}_{i,b}= x_{\pi(i, b)}$ where $\pi(i,b)$ indicate which data point is sampled as the $b$-th instance in the $i$-th batch.
To propagate gradients to synthetic images, we have
\begin{equation}
\frac{\partial \|\hat{\theta}_{t+T}-\theta^*_{t+M}\|_2^2}{\partial x_j} \!=\!\! \sum_{(i,b): \pi(i,b)=j} \!\!\frac{\partial \|\hat{\theta}_{t+T}-\theta^*_{t+M}\|_2^2}{\partial \tilde{X}_{i,b}}.
\label{eq:assignment}
\end{equation}
We can see that \eqref{eq:assignment} is essentially doing gradient accumulation: whenever we obtain the gradient to a batch $\tilde{X}_i$, we accumulate that gradient to $x_j$ if $x_j \in \tilde{X}_i$.
This process does not require storing any computation graphs for the backward pass.
And since we only need to store one accumulated gradient vector for each synthetic image, the memory is also constant w.r.t. $T$.

Since both \eqref{eq:gradient} and \eqref{eq:assignment} do not require maintaining all $T$ computational graphs of $\{ \nabla_{\theta} \ell(\hat{\theta}_{t+i};\tilde{X}_{i})\}_{i=0}^{T-1}$, {\bf the overall memory complexity is constant with respect to $T$}.
 As a result, we can compute the gradient to all the synthetic images with constant memory, at the cost of computing MTT loss consecutively over each batch.
In practice, we found that this consecutive computation incurs negligible runtime overhead, as the total  cost is almost identical to the original MTT computation  (see~\cref{fig:memory_runtime}).

\subsection{Memory complexity v.s. other methods}
\label{sec.complexity}
In this section, we discuss our method's GPU memory usage analytically and compare it with other methods.   
We focus on comparing our method with the original MTT, as well as FrePo, the only existing method that scales to ImageNet-1K under limited IPCs (up to 2). 
We also discuss about other bi-level optimization methods in~\ref{sec.appendix.other_complexity}.

In the following discussion, we use $T$ to denote SGD steps to match trajectories and $X$/$\tilde{X}$ to denote the whole real and synthetic dataset respectively. $\tilde{X}_i \sim \tilde{X}$ then represents a batch of data $\tilde{X}_i$ sampled from entire distilled dataset. 
For simplicity, we further make a moderate approximation that the memory footprint of the computation graph scales linearly w.r.t. the batch size\footnote{This is not strictly the case since some components of the backward graph are independent to batch size, but the scaling law for the rest of the graph is roughly linear.}, and use $\mathcal{G}$ to denote the computation graph size for a single input image.

\textbf{v.s. MTT:} As MTT has to store the computation graphs for the entire matching trajectory, its memory consumption can be written as $\mathcal{O}(T |\tilde{X_i}| \mathcal{G})$ (\cref{eq:square_loss_ref}).
For a predefined batch size $|\tilde{X_i}|$, $T$ increases linearly w.r.t. the dataset size, which significantly limits the MTT's scalability.
In contrast, our method retains a memory complexity of $\mathcal{O}(|\tilde{X}_i|\mathcal{G})$, which is independently of $T$ thanks to the loss decomposition presented in~\cref{eq:gradient}.

\textbf{v.s. FrePo:} We also compare our methods with FrePo - the previous SOTA on ImageNet-1K with IPC 1 and 2. FrePo learns the synthetic images by optimizing the following loss:
\begin{equation}
    \label{eq:frepo}
    \mathcal{L}(\tilde{X}, X) = \frac{1}{2}\| Y_t - K_{X\tilde{X}}^\theta(K_{\tilde{X}\tilde{X}}^\theta + \lambda\textit{I})^{-1}\tilde{Y} \|_2^2
\end{equation}
\begin{equation*}
    K_{X\tilde{X}}^\theta = f(X,\theta)f(\tilde{X},\theta)^T, \hspace{0.5cm} K_{\tilde{X}\tilde{X}}^\theta = f(\tilde{X},\theta)f(\tilde{X}, \theta)^T, 
\end{equation*}
where $f(X, \theta)$ maps $X$ to the latent feature in the last hidden layer of a network parameterized by $\theta$.
Noticably, the second term in~\cref{eq:frepo} is the analytical solution to the inner optimization, hence it uses full batch~\cite{zhou2022dataset}.
It can be seen that FrePo's loss function involves the Gram matrix $K_{\tilde{X}\tilde{X}}^\theta \in \mathbbm{R}^{|\tilde{X}|\times |\tilde{X}|}$, which is computed from feeding all synthetic images into the model.
As a result, FrePo not only incurs quadratic complexity w.r.t. the synthetic dataset size, but also requires storing the computation graphs of the entire synthetic dataset in one pass.
Its overall memory consumption can thus be written as $\mathcal{O}(|\tilde{X}| \mathcal{G}_{frepo} + |\tilde{X}|^2)$\footnote{For a single image, the computation graph of FrePo is slightly smaller than ours since we back-propagate through all layers.}.
For ImageNet-1K with IPC 50, there are $50,000$ synthetic images, which becomes computationally prohibitive to run given its memory complexity.
Moreover, in terms of runtime, FrePo's matrix inversion operation also incurs an extra cubic runtime overhead: $\mathcal{O}(|\tilde{X}|^3)$, whereas our method does not involve any superlinear terms.

\subsection{Soft labels}
\label{sec.soft_label}
Using learned soft labels for synthetic images is a commonly adopted technique in kernel-based distillation methods like FrePo.
Concretely, labels of the synthetic dataset are treated as a learnable parameter that can be jointly optimized with synthetic images.
Compared with one-hot hard labels, soft labels allow information to flow across classes, thereby increasing the compression efficiency.
As a result, it is shown to be critical to the performance of FrePo on datasets with large number of labels such as ImageNet-1K, especially under low IPCs.
For example, FrePo reports 7.5\% test accuracy on ImageNet IPC=1 with soft labels, compared with only 1.6\% using hard labels.

The failure of hard labels can also be observed when scaling matching-based MTT to ImageNet-1K: we found that using hard labels on our memory-efficient MTT also leads to poor results (0.7\% under IPC=1).
However, while kernel-based methods benefit greatly from label learning, it only shows marginal gains in our case (\cref{sec.ablation}).
We conjecture that, although learnable labels bring extra flexibility, updating the labels alongside with synthetic images $\tilde{X}$ and model weight $\hat\theta$ also makes the inner optimization of MTT more challenging to solve.

To unleash the power of soft labels for MTT, we introduce a novel train-free method for assigning soft labels to synthetic images.
Recall that the goal of MTT is to match the parameters of the student model trained on synthetic images to the teacher model trained on real images.
Therefore, we can directly leverage the pre-trained teacher models to generate soft labels.
Concretely, at every iteration, after sampling a trajectory of a teacher model, we pass the synthetic image to the teacher model, store the generated soft labels, and use these labels in the gradient computation of \cref{eq:expansion} for the student model's updates. 
The gradients computed from synthetic images and their soft labels will then be used to form the MTT loss.
Our method can be viewed as a form of knowledge distillation~\cite{hinton2015distilling}, where the knowledge is distilled from the teacher model to the student model through the generated soft labels.
Therefore, it not only helps with learning synthetic images, but also enriches the information condensed into the synthetic dataset.

The proposed Soft Label Assignment (SLA) requires no additional training and extra hyperparameters.
The only design choice is which teacher model checkpoint to use for label assignment.
We discuss two options below:

\textbf{Teacher Model @ Target Step:} Since our method samples a section of the teacher model's trajectory at every iteration, it is natural to use the teacher model at the target matching step (i.e. $\theta_{t+M}^*$) to generate soft labels.
This option is intuitive, as our objective for a single iteration is to match the teacher model at the sampled target step.
Empirically, SLA using target-step teacher model achieves remarkably strong performance, leading to 7\% to 13.4\% absolute accuracy gain on ImageNet-1K across different IPCs.

\textbf{Teacher Model @ Last Epoch:} Since all teacher models are pre-trained prior to optimizing synthetic images, one may wonder whether we can always use the fully-trained teacher models to generate soft labels.
Although a fully-trained teacher model outperforms its intermediate checkpoints, it could also be far away from the sampled trajectory where the matching actually occurs.
As a result, the generated soft labels might not be suitable for guiding the matching process.
Indeed, empirically we found that the performance of SLA using fully-trained teachers is much worse than that of target-step teacher (\cref{fig.soft_hard}).
Therefore, we use the first option for all main experiments. 

The proposed algorithm, {\bf T}raj{\bf E}ctory matching with {\bf S}oft {\bf L}abel {\bf A}ssignment (TESLA), which combines the memory-efficient gradient computation of trajectory matching loss and the soft label assignment method, is summarized in~\cref{alg:algorithm} and~\cref{fig:tesla}.
\begin{algorithm}[tb]
\caption{\textbf{T}raj\textbf{E}ctory matching with \textbf{S}oft \textbf{L}abel \textbf{A}ssignment (TESLA)}
\label{alg:algorithm}
\begin{algorithmic}
\STATE \textbf{Input:} $f:$ teacher model; $\Theta:$ teacher model's trajectories; $K$: number of iterations; $T$: number of matching steps; $\beta$: learning rate for student model; $\alpha$: learning rate for the synthetic images.
\FOR{iter $=1 \dots$ K}
 \STATE Sample $\theta_{t}^*$ and $\theta_{t+M}^*$ $\in\Theta$, set $G=0$, $\hat{\theta}_t=\theta_t^*$
 \STATE Initialize $\tilde{Y}=f(\theta_{t+M}^*;\tilde{X})$ \COMMENT{Soft Label Assignment}
\FOR{$i=1, \dots, T$} 
\STATE Compute $g_i = \nabla_\theta \ell(\hat{\theta}_{t+i};\tilde{X}_i)$
\STATE Update $\hat{\theta}_{t+i} = \hat{\theta}_{t+i-1} - \beta g_i$;\ \  $G = G + g_i$
\ENDFOR
\FOR{$i=1, \dots, T$}
\STATE Compute $ g_i = \nabla_\theta \ell(\hat{\theta}_{t+i};\tilde{X}_i)$
\STATE Compute
$\frac{\partial \|\hat{\theta}_{t+T}-\theta^*_{t+M}\|_2^2}{\partial \tilde{X}_i}$ based on 
$g_i$ and~\cref{eq:gradient}
\STATE $\frac{\partial \|\hat{\theta}_{t+T}-\theta^*_{t+M}\|_2^2}{\partial x_{\pi_{i,b}}} +\!\!=\! \frac{\partial \|\hat{\theta}_{t+T}-\theta^*_{t+M}\|_2^2}{\partial \tilde{X}_{i,b}}$  for all $b$
\ENDFOR
\STATE Update $x_j$ using $\frac{\partial \|\hat{\theta}_{t+T}-\theta^*_{t+M}\|_2^2}{\partial x_{j}}$ for all sampled $j$
\ENDFOR
\end{algorithmic}
\end{algorithm}

\section{Experimental Results}
\subsection{Experiment setup}
\textbf{Experiment Settings:} We evaluate TESLA on 3 datasets including CIFAR-10/100~\cite{krizhevsky2009learning} and ImageNet-1K~\cite{russakovsky2015imagenet} (\cref{sec.appendix.datasets}). On CIFAR-10/100, we follow other methods and learn 1/10/50 image(s) per class. For ImageNet-1K, we resize it to 64$\times$64 resolutions following~\cite{zhou2022dataset}. We learn 10/50 images per class together with 1 and 2 that are reported by previous works. For the surrogate model, we use the same ConvNet architecture as DSA/DM/MTT. The model's convolutional layer consists of 128 filters with kernel size $3\times 3$ followed by Instance normalization\cite{ulyanov2016instance}, RELU activation and an average pooling layer with kernel size $2\times 2$ and stride 2.

Following MTT, we apply ZCA whitening on CIFAR-10/100. On ImageNet-1K, we don't apply any preprocessing techiniques. Simiar to MTT, we apply the same DSA~\cite{radford2015unsupervised, goodfellow2016deep, zhao2020differentiable} augmentation during training and evaluation. When the dataset is simple and doesn't contain many classes such as CIFAR-10/100, soft label is not needed~\cite{zhou2022dataset}. We find soft label most effective on ImageNet-1K. See~\cref{sec.appendix.hyperparameters} for detailed hyperparameters.

\textbf{Evaluation and baselines:} 
Following prior works~\cite{zhaodsa, zhaodm, cazenavette2022dataset, zhou2022dataset, dcbench}, we evaluate the distilled datasets by training five randomly initialized models on them, and report the mean and standard deviation of their accuracy on the real test set. For baseline methods, we directly list numbers from their original paper when they are available.
Since most prior methods do not conduct experiments on ImageNet-1K, we try our best to apply them on ImageNet-1K. Otherwise, we mark them as absent in~\cref{tab:overall_accuracy} and~\cref{table:transfer}. More details can be found in~\cref{sec.appendix.competitors}. 
For KIP, we use their open-sourced dataset to measure the performance since their original work uses a 1024-wide model for evaluation compared to the 128-wide model for other methods and has an extra convolutional layer. FrePo uses a model that doubles the number of filters when the feature map size is halved while other works use the same number of filters for all convolutional layers~\cite{zhaodsa, zhaodm, cazenavette2022dataset}, thus the model used by FrePo has a lot more parameters\footnote{22.6M  trainable parameters from FrePo compared to 2.5M trainable parameters from other methods on a 4-layer ConvNet. } than other methods. We still report FrePo's original results due to the lack of open-sourced code and publicly available dataset. 
\vspace{-5pt}
\subsection{Empirical results}
\vspace{-5pt}
We compare TESLA against the random baseline and previous SOTA methods including DSA~\cite{zhaodsa}, DM~\cite{zhaodm}, KIP~\cite{nguyen2021dataset}, FrePo~\cite{zhou2022dataset} and the original MTT. The results are presented in~\cref{tab:overall_accuracy}. 
On smaller datasets, our method outperforms prior arts with the same model architecture.
On ImageNet-1K, TESLA outperforms FrePo and DM with IPC 1 and 2. On 10 and 50 IPCs where all existing methods fail to scale, {TESLA only incurs 16\% and 5.9\% accuracy drop compared to training with the whole ImageNet-1K dataset while using only 0.83\% and 4.2\% of its training dataset size. This is an 18.2\% accuracy improvement  to the prior art~\cite{zhou2022dataset}}.
\begin{table*}[ht]
\centering
\caption{Test accuracies of models trained on synthetic dataset.}
\resizebox{1.0\textwidth}{!}{
  \begin{threeparttable}
\begin{tabular}{cccccccccc}
\toprule
Dataset & IPC & Random & DSA & DM & KIP\tnote{1} & FrePo\tnote{2} & MTT & \textbf{TESLA(Ours)}\tnote{3} & Whole Dataset\\
\\ \toprule
\multirow{3}{*}{CIFAR10} & 1 & 15.4$\pm$0.3 & 36.7$\pm$0.8 & 31.0$\pm$0.6 & 40.6$\pm$1.0 (\textbf{49.9$\pm$0.2}) & 46.8$\pm$0.7 & 46.3$\pm0.8$ & \textbf{48.5$\pm$0.8} & \multirow{3}{*}{86.0$\pm$0.1} \\
 & 10 & 31.0$\pm$0.5 & 53.2$\pm$0.8 & 49.2$\pm$0.8 & 47.2$\pm$0.7 (62.7$\pm$0.3) & 65.5~$\pm$0.6 & 65.3$\pm0.7$ & \textbf{66.4$\pm$0.8}\\
 & 50 & 50.6$\pm$0.3 & 66.8$\pm$0.4 & 63.7$\pm$0.5 & 57.0$\pm$0.4 (68.6$\pm$ 0.2) & 71.7$\pm$0.2 & 71.6$\pm$0.2 & \textbf{72.6$\pm$0.7}\\
\midrule
\multirow{3}{*}{CIFAR100} & 1 & 5.3$\pm$0.2 & 16.8$\pm$0.2 & 12.2$\pm$0.4 & 12.0$\pm$0.2 (15.7$\pm$0.2)\tnote{*} & \textbf{27.2$\pm$0.4}\tnote{*} & 24.3$\pm$0.3 & \textbf{24.8$\pm$0.4} & \multirow{3}{*}{56.7$\pm$0.2}\\
& 10 & 18.6$\pm$0.25 & 32.3$\pm$0.3 & 29.7$\pm$0.3 & 29.0$\pm$0.3 (28.3$\pm$0.1) & 41.3$\pm$0.2\tnote{*} & 40.6$\pm$0.4 & \textbf{41.7$\pm$0.3}\\
& 50 & 34.7$\pm$0.4 & 42.8$\pm$0.4 & 43.6$\pm$0.4 & - & 44.3$\pm$0.2\tnote{*} & 47.7$\pm$0.2 & \textbf{47.9$\pm$0.3}\\
\midrule
\multirow{4}{*}{ImageNet-1K} & 1 & 0.5$\pm$0.1  & - & 1.5$\pm$0.1 & - & 7.5$\pm$0.3\tnote{*} & - & \textbf{7.7$\pm$0.2}\tnote{*} &\multirow{4}{*}{33.8$\pm$0.3} \\
&2&0.9$\pm$0.1 & -  & 1.7$\pm$0.1 & - & 9.7$\pm$0.2\tnote{*} & - & \textbf{10.5$\pm$0.3}\tnote{*}\\
&10& 3.6$\pm$0.1& -& - & -& - & -  & \textbf{17.8$\pm$1.3}\tnote{*}\\
&50& 15.3$\pm$2.3& -& -& -& - & - & \textbf{27.9$\pm$1.2}\tnote{*}\\
\bottomrule
\end{tabular}
  \begin{tablenotes}
    \item [*] Soft labels are used when table entries are marked with *.
    \item[1] KIP's performance is measured with the dataset released by the author. Performances in quotas are from the original paper under different settings.
    \item [2]FrePo uses a different model with much more parameters. We still mark FrePo result as bold if it outperforms other methods.
    \item[3] Our performances are achieved using slightly different hyperparameters than MTT, see~\cref{sec.appendix.hyperparameters}.
    \item Entries marked as absent are due to scability issues. See~\cref{sec.appendix.competitors} for detailed reasons.
    \end{tablenotes}
\end{threeparttable}
}
\label{tab:overall_accuracy}
\end{table*}
\vspace{-5pt}
\subsection{Training cost analysis}
\label{sec.cost}
\vspace{-5pt}
As discussed in~\cref{sec.constant_memory}, a key benefit of our method over MTT is constant memory consumption w.r.t. the matching steps, with only marginal runtime overhead. In this section, we empirically benchmark and compare the memory and runtime of our methods against MTT\footnote{FrePo is only compared analytically due to lack of open-source code.}.

\begin{figure}[ht]
\centering
\includegraphics[width=0.9\columnwidth]{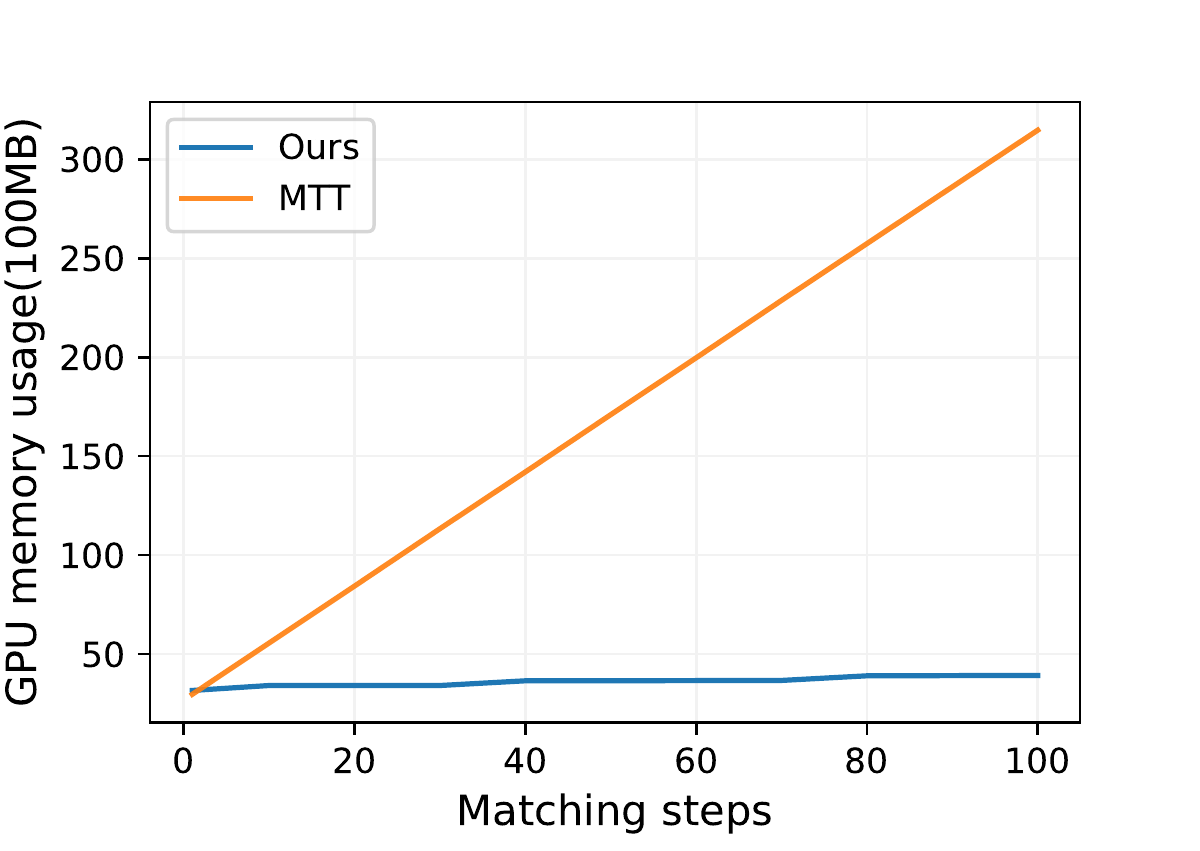}
\vspace{-10pt}
\caption{GPU memory usage comparison between MTT and TESLA. MTT scales linearly w.r.t. the number of matching steps, whereas our method uses constant memory. Results are measured on CIFAR100 with batch size 100 under varying matching steps.}
\vspace{-10pt}
\label{fig:memory_cost}
\end{figure}

We first compare the GPU memory consumption between our method and MTT. For this experiment, we keep everything else the same between two methods, and only vary the matching steps.
The results are shown in~\cref{fig:memory_cost} (The numerical results can be found in~\cref{tab:cost}).
The memory consumption of the original MTT increases linearly with the number of synthetic steps, while it remains constant for our method.
This observation aligns with our theoretical analysis in~\cref{sec.complexity}.
In principle, the constant memory reduction allows us to scale to arbitrarily large IPCs.
\begin{figure}[ht]
\centering
    \includegraphics[width=0.9\columnwidth]{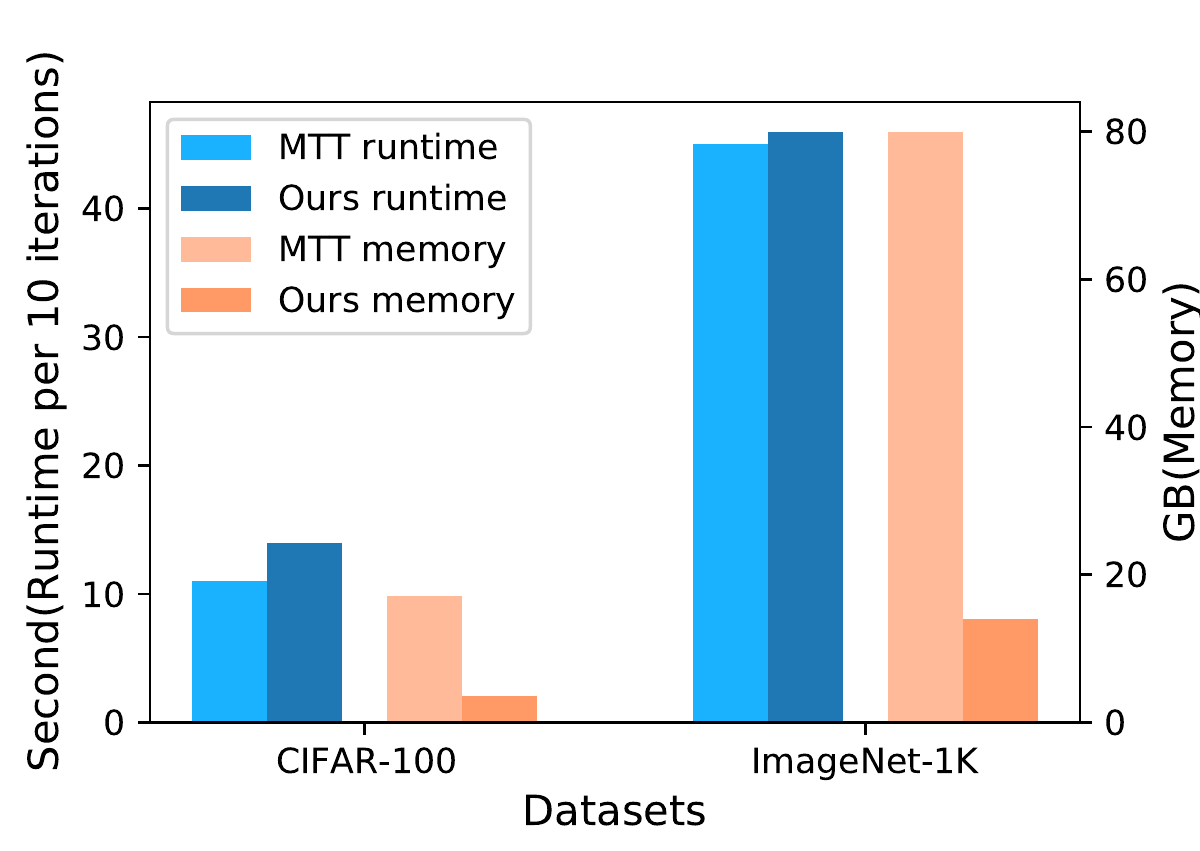}
    \vspace{-10pt}
    \caption{GPU memory and runtime comparison between MTT and TESLA on different datasets. Results are measured with a batch size of 100 and 50 matching steps.}
    \label{fig:memory_runtime}
\vspace{-10pt}
\end{figure}
We proceed to test the runtime overhead, alongside memory consumption across different dataset\footnote{We don't measure it on CIFAR-10 because the synthetic dataset is too small even with IPC 50}.
For this experiment, we fix the synthetic training step to 50 and batch size to 100.
The results are summarized in~\cref{fig:memory_runtime} (See~\cref{tab:memory_runtime} for numerical results).
On CIFAR-100, our method obtains $\sim 5$x memory reduction over MTT, while only introduces $\sim27\%$ overhead runtime. \textbf{On ImageNet-1K, TESLA obtains $\bf{\sim 6}$x memory reduction with only  $\bf{\sim 2}$\% extra time\footnote{MTT's runtime on ImageNet-1K is estimated since MTT is OOM under our settings. See~\cref{sec.appendix.cost}} compared to MTT}.

\begin{table*}[ht]
\centering
\caption{Test accuracy on ConvNet versus transferred to other architectures. All methods are evaluated with 10 IPCs.}
\resizebox{0.85\textwidth}{!}{
\begin{threeparttable} 
\begin{tabular}{c|cccccccccccccc}
\toprule
&\multicolumn{3}{c}{CIFAR-10}
&\multicolumn{3}{c}{CIFAR-100} & 
\multicolumn{3}{c}{ImageNet-1K}\\
& ConvNet & ResNet18 & ViT & ConvNet &  ResNet18 & ViT & ConvNet & ResNet18 & ViT  \\
\midrule
Random  & 31.0$\pm$0.5 &29.6$\pm$0.9 &26.2$\pm$0.5 &18.6$\pm$0.3 & 15.8$\pm$0.2 & 14.1$\pm$0.2 &3.6$\pm$0.1 &1.4$\pm$0.1&3.2$\pm$0.0\\
DSA  & 53.0$\pm$0.4&42.1$\pm$0.6&31.9$\pm$0.4 & 32.2$\pm$0.4 & 21.9$\pm$0.4 & 19.6$\pm$0.2 & - & - & -  \\
DM &47.6$\pm$0.6 &38.2$\pm$1.1 &34.4$\pm$0.5 & 29.2$\pm$0.3 & 18.7$\pm$0.5 & 17.1$\pm$0.3 & - & - & -  \\
KIP &47.2$\pm$0.4 &38.8$\pm$0.7 & 15.9 $\pm$1.1 & 29.0$\pm$0.3 & 20.1$\pm$0.5 & 12.1$\pm$0.7& - & - & -  \\
MTT & 65.3$\pm$0.7 &46.1$\pm$1.4 &34.6$\pm$0.6 & 40.6$\pm$0.4 & 26.8$\pm$0.6 & 20.4$\pm$0.2 & - & - & -  \\
\textbf{Ours} & \textbf{66.4$\pm$0.8} & \textbf{48.9$\pm$2.2}& \textbf{34.8$\pm$1.2}& \textbf{41.7$\pm$0.3} & \textbf{27.1$\pm$0.7} & \textbf{21.0$\pm$0.3} & \textbf{17.8$\pm$1.3} & \textbf{7.7$\pm$0.1} & \textbf{11.0$\pm$0.2}\\
\bottomrule
\end{tabular}
\end{threeparttable}
}
\label{table:transfer}
\vspace{-10pt}\end{table*}

\subsection{Ablation study on soft labels}
\label{sec.ablation}
We conduct two ablation studies on ImageNet-1K to compare the effectiveness of soft labels.
First, we study our method with soft labels and hard labels and show the results in~\cref{table:hard_soft}. Our method with soft labels outperforms hard labels by a large margin, e.g. 7\% on IPC 1 and 13.4\% on IPC 10, showing the effectiveness of soft labels.
We proceed to investigate several other soft label strategies as follows.

\begin{figure}[ht]
\vskip -0.2in
\centering
\includegraphics[width=0.9\columnwidth]{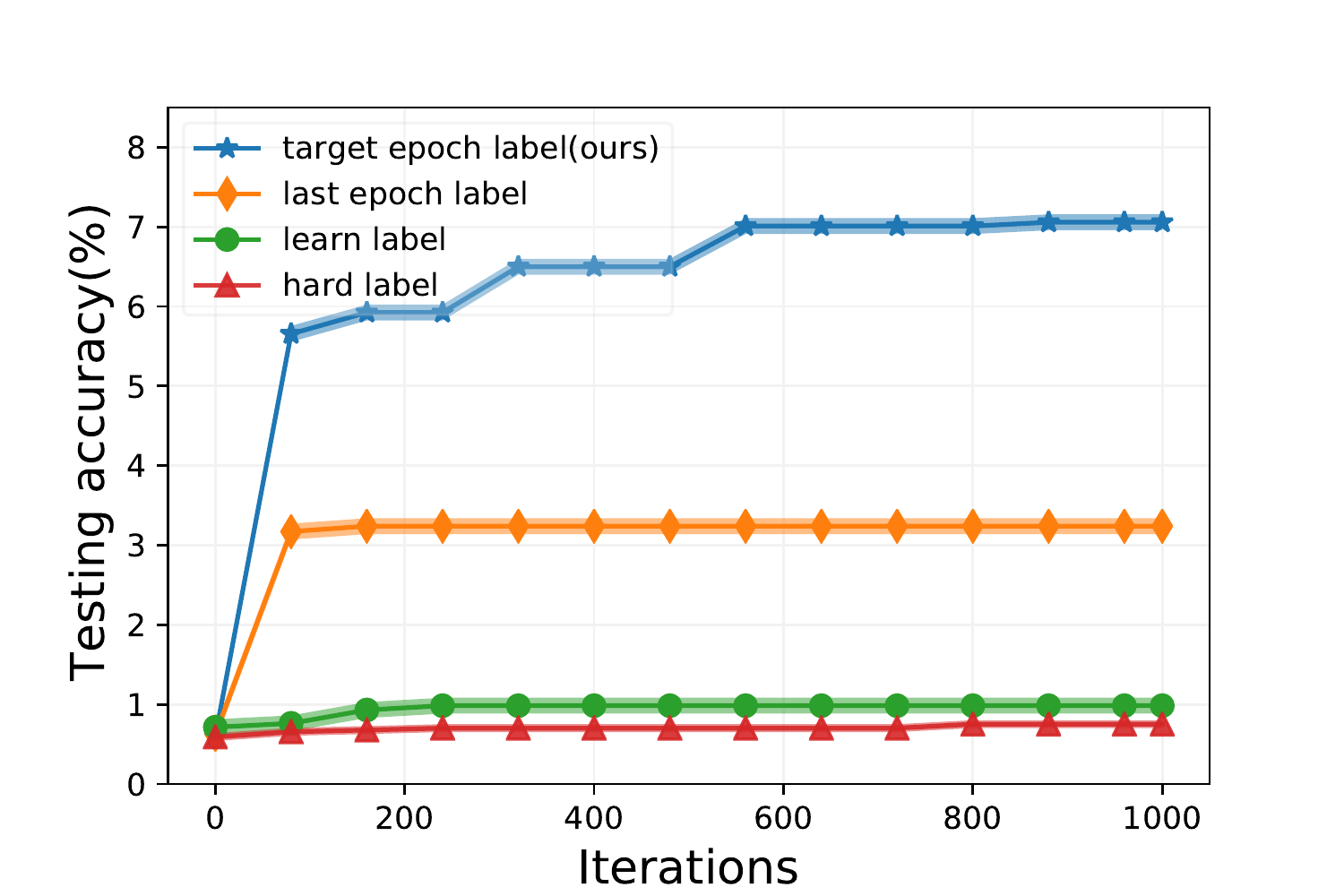}
  \caption{Ablation study on different label strategies.  Y-axis shows the maximum accuracy achieved until that iteration.}
  \label{fig.soft_hard}
\vskip -0.2in
\end{figure}

\textbf{Label Learning:} In this experiment, we study the strategy of learning labels instead of generating them from teacher models. We initialize the pre-softmax logits so that the probability after softmax is close to one-hot (\cref{sec.appendix.label_learning}).
The results are plotted in~\cref{fig.soft_hard}.
While learning labels do slightly improve the performance, the margin of gain is far less  compared with those reported on kernel-based methods such as FrePo and KIP.
The algorithm still fails to update the synthetic dataset effectively, even with the extra flexibility of the learned labels.
Note that we also experiment with different label learning strategies, such as directly initializing and optimizing post-softmax labels (hence allowing each label to move beyond 0-1 range), but the results are similar.
\begin{table}
\centering
  \caption{Ablation study on testing accuracy (\%) using hard vs soft label on ImageNet-1K at 1500 iterations.}
  \resizebox{\columnwidth}{!}{
  \begin{threeparttable}
  \begin{tabular}{ccccc}
    \toprule
    IPC & 1 & 2& 10 & 50 \\
    \midrule
    Hard label & 0.7$\pm$0.1 & 1.1$\pm$0.1& 4.4$\pm$0.3 & 18.1$\pm$1.5\\
    TESLA & 7.7$\pm$0.2 &10.5$\pm$0.3 & 17.8$\pm$1.3 & 27.9$\pm$1.2\\
    \bottomrule
  \end{tabular}
  \end{threeparttable}
  }
  \label{table:hard_soft}
\vspace{-10pt}
\end{table}

\textbf{Target (Ours) vs Last Epoch:} We also study soft labels generated by the teacher model 
 at the target step versus the last epoch. It's natural to think that a better-trained model will capture more training data statistics, thus generating better soft labels. However, we find that this doesn't work with trajectory matching. As shown in~\cref{fig.soft_hard}, the algorithm fails to learn effectively with last epoch parameters.

\begin{figure}[ht]
\centering
\includegraphics[width=0.9\columnwidth]{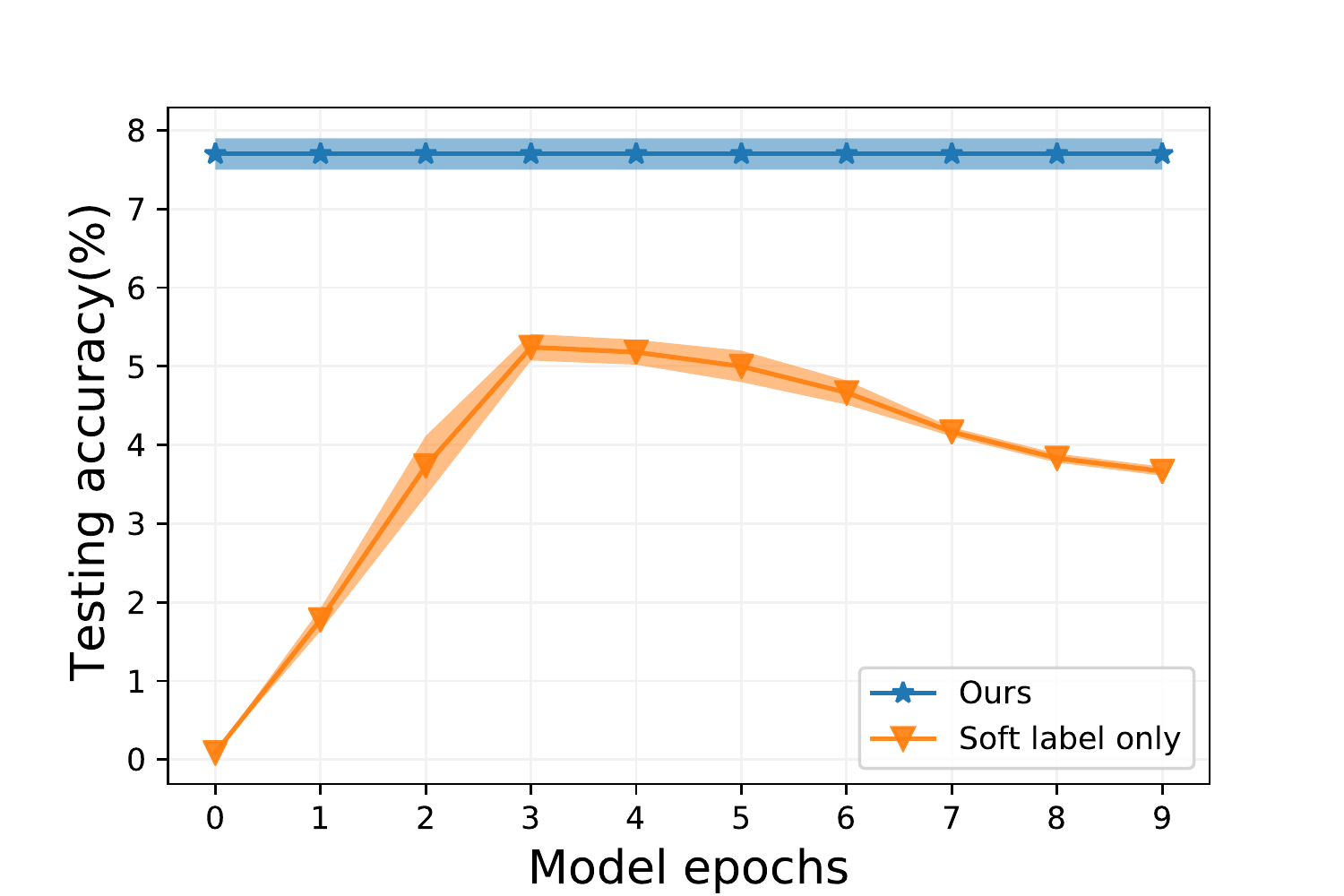}\centering
\vspace{-10pt}
  \caption{Ablation study on performance of SLA without updating synthetic images. Top flat line shows the performance of TESLA baseline.}
  \vspace{-10pt}
  \label{fig.soft_label_only}
\vskip -0.2in
\end{figure}

We found that soft label assignment benefits synthetic image learning. To show this, we study the effect of soft label assignment technique alone, by fixing the synthetic images($\tilde{X}$). Then we measure the impact of soft labels($\tilde{Y}$) produced by the teacher model with parameter $\theta_{t+M}^*$.
On ImageNet-1K IPC 1, we achieve state-of-the-art performances by iteratively setting $\theta_{t+M}^*$ as parameters from one of the first 9 epochs\footnote{Same as MTT, we always sample $\theta_{t+M}^*$ from teacher trajectories after a full epoch. One epoch contains multiple SGD steps} of the teacher model (SLA step in~\cref{alg:algorithm}). In the ablation study, we randomly select 1 image per class and generate their labels using teacher models from epoch 0 to epoch 9. The results are shown in~\cref{fig.soft_label_only}. It can be seen that around 5.3\% accuracy can be achieved by initializing the labels using teacher models without updating synthetic images. And our method is able to achieve around 7.7\% testing accuracy by integrating soft labels with our memory-efficient implementation of MTT.

\subsection{Cross-Architecture generalization}
Following previous works~\cite{zhaodm,cazenavette2022dataset,zhou2022dataset, dcbench}, we evaluate the transferability of our condensed dataset in training new architectures unseen in the synthetic dataset generation phase. The experiment is conducted on CIFAR-10, CIFAR-100 and ImageNet-1K under 10 IPCs.
Besides the baseline vanilla ConvNet model, we report performance on ResNet18 and ViT~\cite{dosovitskiy2020image, dcbench}.
As shown in~\cref{table:transfer}, our method transfers well across datasets and models, outperforming previous methods by a sizable margin.
This shows that the proposed method can be empirically effective in distilling generalizable information into the synthetic dataset. We are not able to get FrePo's performances due to the lack of open-sourced code and publicly available distilled dataset.

\section{Conclusion}
We present a novel method to reduce the previous SOTA: MTT's heavy memory cost from $\mathcal{O}(T)$ to $\mathcal{O}(1)$ with negligible run-time overhead. We also propose soft label assignment to guide the matching process of model training trajectories. By combining the two, we are able to scale dataset distillation to ImageNet-1K with IPC 10 and 50 for the first time, achieving SOTA performance.
Moverover, our distilled dataset transfer well to across different architectures, such as ViT.
We hope our method can pave the way for future works to explore and expand dataset distillation methods on large-scale real-world datasets.

{\bf Limitation} Despite having constant memory cost with respect to the number of unrolled steps, our method still requires storing some checkpoints of the teacher model, similar to the original MTT. Removing the requirement of a teacher model will be an interesting future research direction to pursue. Further, the study of dataset distillation is in the very early stage so it remains important to close the performance gap to the full model.

\section*{Acknowledgements}
The authors thank the support from NSF  IIS-$2008173$, IIS-$2048280$ and Cisco research award. 


\nocite{langley00}

\bibliography{main}
\bibliographystyle{icml2023}

\newpage
\appendix
\onecolumn
\section{Appendix}

\subsection{Bi-level optimization}
\label{bi-level}
Dataset distillation is originally done through bi-level optimization\cite{wang2018dataset,zhaodc,zhaodsa, cazenavette2022dataset}. Suppose we have a dataset $\mathcal{D}=\{X,Y\}$, the goal of dataset distillation is to get a subset $\mathcal{S}=\{\hat{X}, \hat{Y}\}$ where $|\mathcal{S}| \ll |\mathcal{D}|$ such that the following goals can be achieved $\mathbf{E}_{(x,y)\sim P_\mathcal{D}}\mathcal{L}_{\theta}(x, y) \simeq \mathbf{E}_{(\hat{x}, \hat{y})\sim P_{\mathcal{S}}}\mathcal{L}_{\hat{\theta}}(\hat{x}, \hat{y})$
. Here $\mathcal{L}$ is a loss function. $\theta$ and $\hat{\theta}$ are the parameters generated by using the original dataset and the distilled dataset. It can naturally be divided into a bi-level optimization process where the inner loop updates $\hat{\theta}=\underset{\hat{\theta}}{argmin}\mathcal{L}_{\hat{\theta}}(\hat{x}, \hat{y})$ and the outer loop computes $\mathcal{S}^*=\underset{\mathcal{S}}{argmin}\mathcal{M}_{\hat{\theta}}(\mathcal{S}, \mathcal{D})$. Here $\mathcal{M}$ is a surrogate function that's aimed to identify the  difference between $\mathcal{S}$ and $\mathcal{D}$. For example, \cite{zhaodc} tries to match the training gradients generated by using $\mathcal{S}$ and $\mathcal{D}$, in \cite{cazenavette2022dataset}, the training trajectories are used.

\subsection{Complexity of other distillation methods}
\label{sec.appendix.other_complexity}
In the main text, we focus on the complexity analysis of FrePo and MTT. Here we discuss the complexity of other methods. \cite{wang2018dataset, zhaodc, zhaodsa} also uses bi-level optimization. However, compared to MTT and FrePo, they achieve lower performances and cannot scale to large datasets.  \cite{zhaodm} takes another approach by matching the distributions of synthetic dataset with real dataset. The method is fast due to its single level of optimization. However, it suffers from great accuracy loss when applying to large datasets~(Table~\ref{tab:overall_accuracy}).

\subsection{Datasets}
\label{sec.appendix.datasets}
On 3 datasets including CIFAR-10, CIFAR-100~\cite{krizhevsky2009learning} and ImageNet-1K~\cite{russakovsky2015imagenet}. CIFAR-10/100 includes 50,000 training and 10,000 testing images in 32$\times$32 resolution from 10 and 100 classes respectively. ImageNet-1K is a standard large-scale dataset consists of 1,000 classes with 1,281,167 training and 50,000 testing images. We resize ImageNet-1K images to 64$\times$64 resolutions following~\cite{zhou2022dataset}.
We do not evaluate our methods on toy datasets such as MNIST~\cite{lecun1998gradient} or Fashion-MNIST\cite{xiao2017fashion} as the performances of different methods on MNIST-variants are too close and we target large datasets. 

\subsection{Data preprocessing}
ZCA whitening is first used in KIP. We reimplement the author's custom ZCA in order to evaluate its performance through the publicly released dataset. However, for the ZCA preprocessing in our work, we follow the same Kornia~\cite{EdgarRiba2019KorniaAO} ZCA as the original MTT work. We also follow the same ZCA settings as MTT. On ImageNet-1K, we don't apply ZCA at any of the IPCs.

\subsection{Models}
we use the same ConvNet architecture as DSA/DM/MTT. The model's convolutional layer consists of 128 filters with kernel size $3\times 3$ followed by instance normalization\cite{ulyanov2016instance}, RELU activation and an average pooling layer with kernel size $2\times 2$ and stride 2. The original KIP uses a larger model with a width of 1024 for evaluation compared to 128 used by other methods and it has one more conv layer than the model used by other methods. FrePo uses another different model that doubles the number of filters when the feature map size is halved. Also batch normalization is used by FrePo instead of instance normalization. A 3-layer ConvNet is used for CIFAR-10/100 and a 4-layer ConvNet is used for ImageNet-1K.

\subsection{Hardwares}
All of our experiments are run using one NVIDIA A6000 GPU with 49GB of memory. When measuring the memory consumption used by the original MTT, if it doesn't fit into one GPU, we use two NVIDIA A6000 GPUs just for measuring purpose.

\begin{figure}
    \centering
    \includegraphics[width=\linewidth]{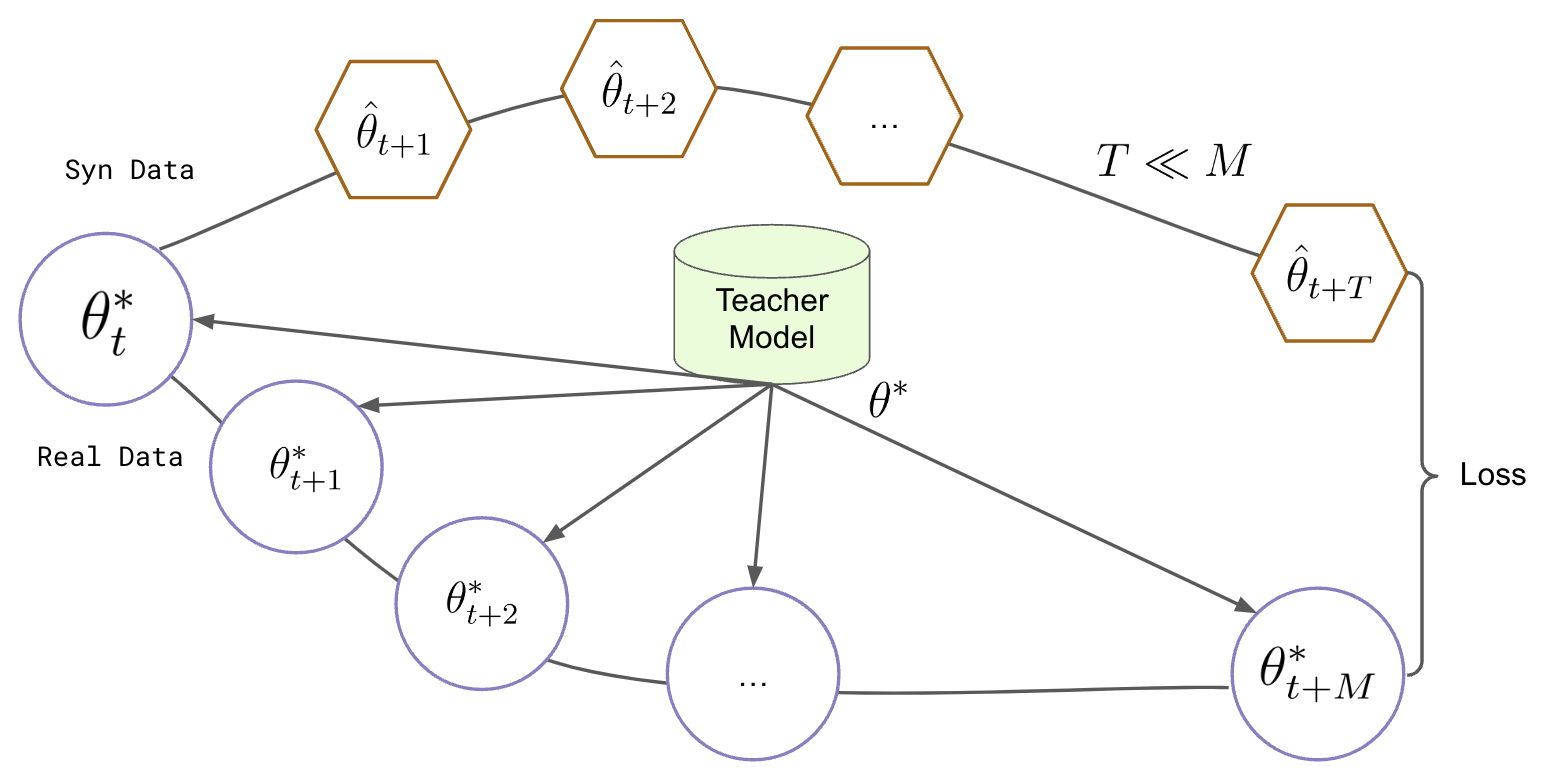}
    \caption{Illustration of matching training trajectories(\textbf{MTT}).}
    \label{fig:mtt}
\end{figure}

\subsection{Label learning}
\label{sec.appendix.label_learning}
In our experiment with label learning, we consider the following 2 implementation options. Firstly we try to initialize labels using one-hot float vectors. Then we directly feed the one-hot vector to the surrogate model and use back-propagation to update the labels. In this scenario, we observe that the classes probabilities will sum up to an arbitrary number after being updated. This causes the cross-entropy loss function used by the surrogate model to be unstable. We observe poor performance by following this implementation.

The second implementation is used in our experiment in~\cref{fig.soft_hard}. Instead of initializing the labels with class probabilities that sum up to 1, we initialize it with logits. In our case, we set the class index bit to 10 and the rest to 0. Then before feeding it into the loss function, we apply a softmax operation to convert logits to class probabilities. The algorithm now is able to learn stably. And we noticed the class weight shift from the class index bit to other classes. However, as mentioned in the main text that it's still not able to learn effectively because of the large label space.

We also consider using different loss functions other than cross entropy. For example, FrePo uses MSE loss in all its experiments. However, we also notice poor performances. As mentioned in~\cite{zhou2022dataset}, cross-entropy loss usually outperforms MSE loss which aligns with our observations.

\subsection{Training cost analysis}
\label{sec.appendix.cost}
We show the runtime and memory analytically in the main text in ~\cref{sec.complexity} and empirically in~\cref{fig:memory_cost} and~\cref{fig:memory_runtime}. Here we include the numerical results in~\cref{tab:cost} and~\cref{tab:memory_runtime}. Note that on ImageNet-1K, the memory usage of MTT is acquired by running it on 2 NVIDIA RTX A6000 GPUs, each of which has 49GB of memory. For MTT's runtime on ImageNet-1K, it's estimated by observing linearity between batch size and runtime.  It can be easily seen that our memory cost is constant with respect to the number of synthetic steps. From~\cref{tab:memory_runtime} we can see that, our algorithm uses only one fifth of MTT's memory with only 27\% more time\footnote{Although our method only computes the first order gradient twice, it's fast and memory efficient compared to the meta gradient computation w.r.t synthetic images which is only computed once.}. Thus our algorithm is able to easily scale to larger datasets with large IPCs.

\subsection{Augmentation}
We use the same DSA augmentation as other works~\cite{zhaodsa, cazenavette2022dataset, zhou2022dataset}. Unlike FrePo which only applies augmentation at evaluation stage, we apply data augmentation at both training and evaluation stage. Similar to previous work such as MTT, we observe better performances. We also try different augmentations such as Autoaugment~\cite{cubuk2018autoaugment} and Randaugment~\cite{cubuk2020randaugment} besides DSA during evaluation, however, we usually observe downgraded performs.

\subsection{Competitors}
\label{sec.appendix.competitors}
In~\cref{tab:overall_accuracy}, we ar not able to get the performance of some methods. We list the reasons here. For DSA, we are not able to get the performances on ImageNet-1K because of memory constraints. For DM, we are only able to get the performances for IPC 1 and 2. We have to perform early stopping when the testing accuracy plateaus because it takes 5 minutes to perform each iteration and the original number of iterations is set to 20,000 by the author. DM gets OOM error when setting IPC to 10 on ImageNet-1K. For KIP, the work is not open-sourced. The author only released distilled datasets on CIFAR-10 with IPC 1,10, 50 and CIFAR-100 with IPC 1 and 10. For FrePo, as it's a recent work, there is no open-sourced code or publicly released dataset yet. Therefore, we report the numbers from its original work. Also as reported by the authors, FrePo is not able to scale to TinyImageNet with IPC 50 and ImageNet-1K with IPC 10.

\subsection{Learning learning rate}
We also notice that learning the student model learning rate is also extremely helpful in generating the synthetic datasets. 
\begin{figure}
    \centering
    \includegraphics[width=0.4\textwidth]{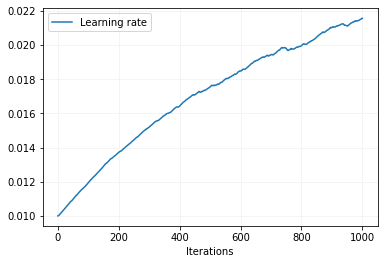}
    \caption{Learning curve for student model learning rate}
    \label{fig:learning_rate}
\end{figure}
We show an example figure of learning rate change for ImageNet-1K IPC 1. It can be seen from~\cref{fig:learning_rate} that as the training progresses, the learning rate will keep increasing. Under our settings, we set its initial values to 0.01. After 1000 iterations, the learning rate goes up to 0.02. We notice that this not only stabilizes the training process, but also improves the testing accuracy when evaluating the distilled datasets.

\begin{table}
\begin{center}
\caption{Peak memory usage comparison on CIFAR-100 between MTT and ours using batch size 100. The units are in MB. \textit{Matching step} means how many gradient descent steps to run before performing model training trajectory matching.}
\resizebox{1\linewidth}{!}{
\begin{threeparttable}
\begin{tabular}{c|ccccccccccc}
\toprule
& \multicolumn{11}{c}{matching step}\\
& 1 & 10 & 20 & 30 & 40 & 50 & 60 & 70 & 80 & 90 & 100\\
\midrule
MTT & 2961 & 5555 & 8431 & 11329 & 14197 & 17093 & 19965 & 22867 & 25737 & 28607 & 31469\\
\midrule
Ours & 3155 & 3405 & 3405 & 3405 & 3643 & 3647 & 3653 & 3661 & 3895 & 3905 & 3913\\
\bottomrule
\end{tabular}
\begin{tablenotes}
  \item The results are measured using one NVIDIA A6000 GPU.
\end{tablenotes}
\end{threeparttable}
}
\end{center}
\label{tab:cost}
\end{table}

\begin{table}
\begin{center}
\caption{Memory and runtime for MTT and ours. The results are measured using batch size 100 and 50 matching steps. The memory is the peak memory used and the runtime is measured for 10 iterations(one iteration includes 50 matching steps). Memory diff is calculated by dividing MTT memory by our memory. Time diff is calculated by dividing our time by MTT time.}
\resizebox{1\linewidth}{!}{
\begin{threeparttable}
\begin{tabular}{c|cccccc}
\toprule
& MTT memory & Our memory & memory diff & MTT time & Our time & time diff \\
\midrule
CIFAR-100 & 17.1$\pm$0.1GB & 3.6$\pm$0.1GB & 4.75X & 11.5$\pm$0.5sec & 14.5$\pm$0.5sec & 1.25X\\
ImageNet-1K & 79.9$\pm$0.1GB & 13.9$\pm$0.1GB & 5.75X & 45.0$\pm$0.4sec & 46.0$\pm$0.8sec & 1.02X \\
\bottomrule
\end{tabular}
\begin{tablenotes}
  \item The results are measured using one NVIDIA A6000 GPU.
\end{tablenotes}
\end{threeparttable}
}
\end{center}
\label{tab:memory_runtime}
\end{table}

\subsection{Hyperparameters}
\label{sec.appendix.hyperparameters}
To facilitate the reproduce of our work, we list all the hyperparameters used in our experiments in this table. We use slightly different hyperparameters on CIFAR-10/100  as listed in~\cref{tab:appendix:hyperparameters} than the original MTT while keeping everything else the same such as model architectures and augmentations. Theoretically, MTT should have the same performances as us on CIFAR-10/100 with smaller IPCs where batch is not needed.
\begin{table}[H]
\begin{center}
\caption{Hyperparameters used to get the distilled dataset.}
\resizebox{1\linewidth}{!}{
\begin{threeparttable}
\begin{tabular}{ccc|cccc|c}
\toprule
Dataset & Model & IPC & Matching Steps & Teacher Epochs & Max Start Epoch & Batch Size & ZCA \\
\midrule
\multirow{3}*{CIFAR-10} & \multirow{3}*{ConvNetD3} & 1 & 50 & 2 & 3 & - & Y \\
& & 10 & 30 & 2 & 20 & - & Y \\
& & 50 & 30 & 3 & 40 & - & N \\
\midrule
\multirow{3}*{CIFAR-100} & \multirow{3}*{ConvNetD3} & 1 & 20 & 3 & 20 & - & Y\\
& & 10 & 15 & 3 & 30 & - & N \\
& & 50 & 50 & 2 & 40 & 100 & Y\\
\midrule
\multirow{4}*{ImageNet-1K} & \multirow{4}*{ConvNetD4} & 1 & 10 & 3 & 6 & 100 & N\\
& & 2 & 15 & 3 & 10 & 100 & N \\
& & 10 & 20 & 3 & 10 & 500 & N\\
& & 50 & 100 & 3 & 25 & 500 & N \\
\bottomrule
\end{tabular}
\end{threeparttable}
}
\end{center}
\label{tab:appendix:hyperparameters}
\end{table}

\subsection{Example distilled image}
We show example distilled images for the 3 dataset used in this work for easier references. For CIFAR-10, we show 10 images per class, for CIFAR-100, we show 1 image per class and for ImageNet-1K, we show 1 image per class for the 1K classes.
(more pages after this paragraph)

\newpage
\begin{figure*}
    \centering
    \includegraphics[width=\textwidth]{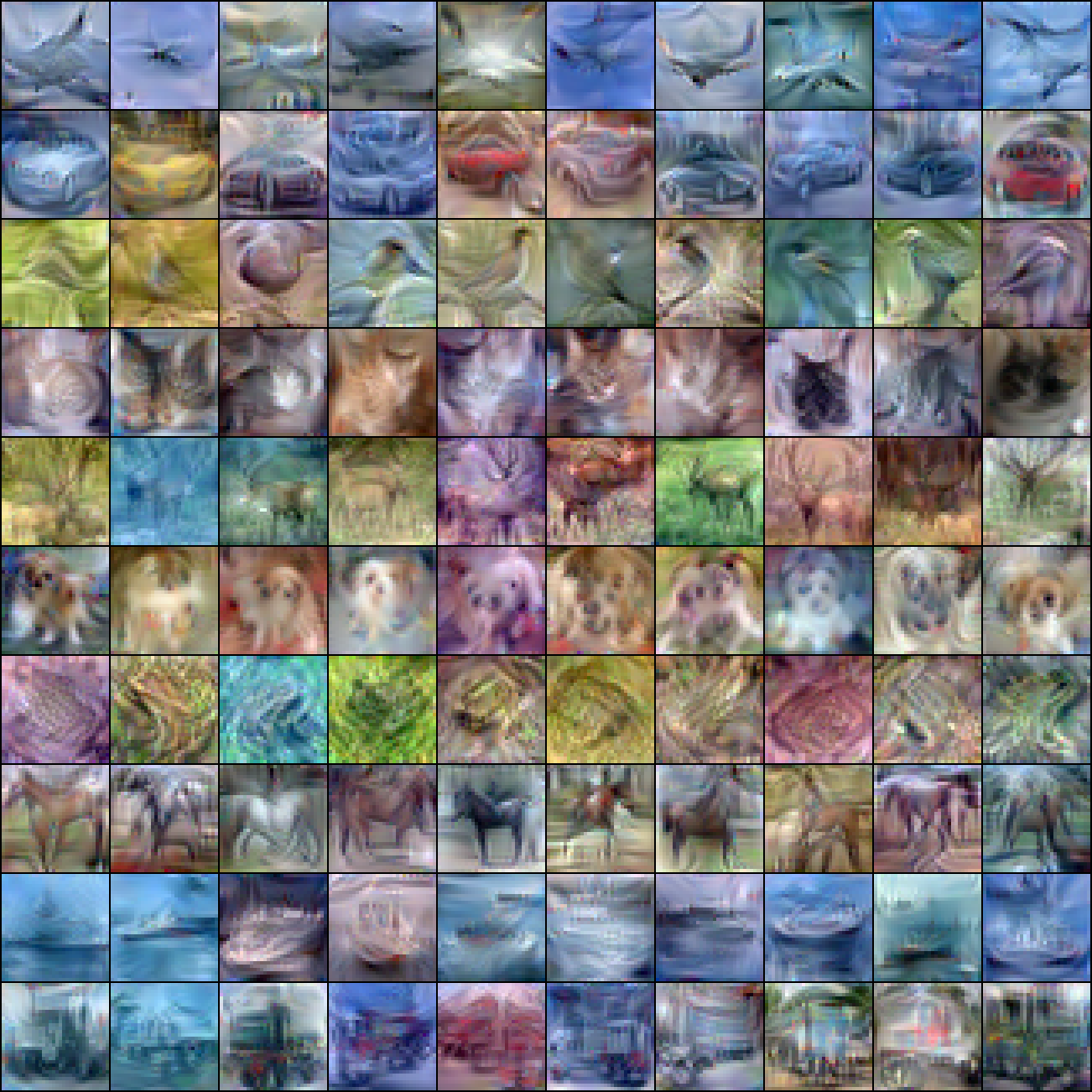}
    \caption{CIFAR10 IPC 10}
\end{figure*}
\begin{figure*}
    \centering
    \includegraphics[width=\textwidth]{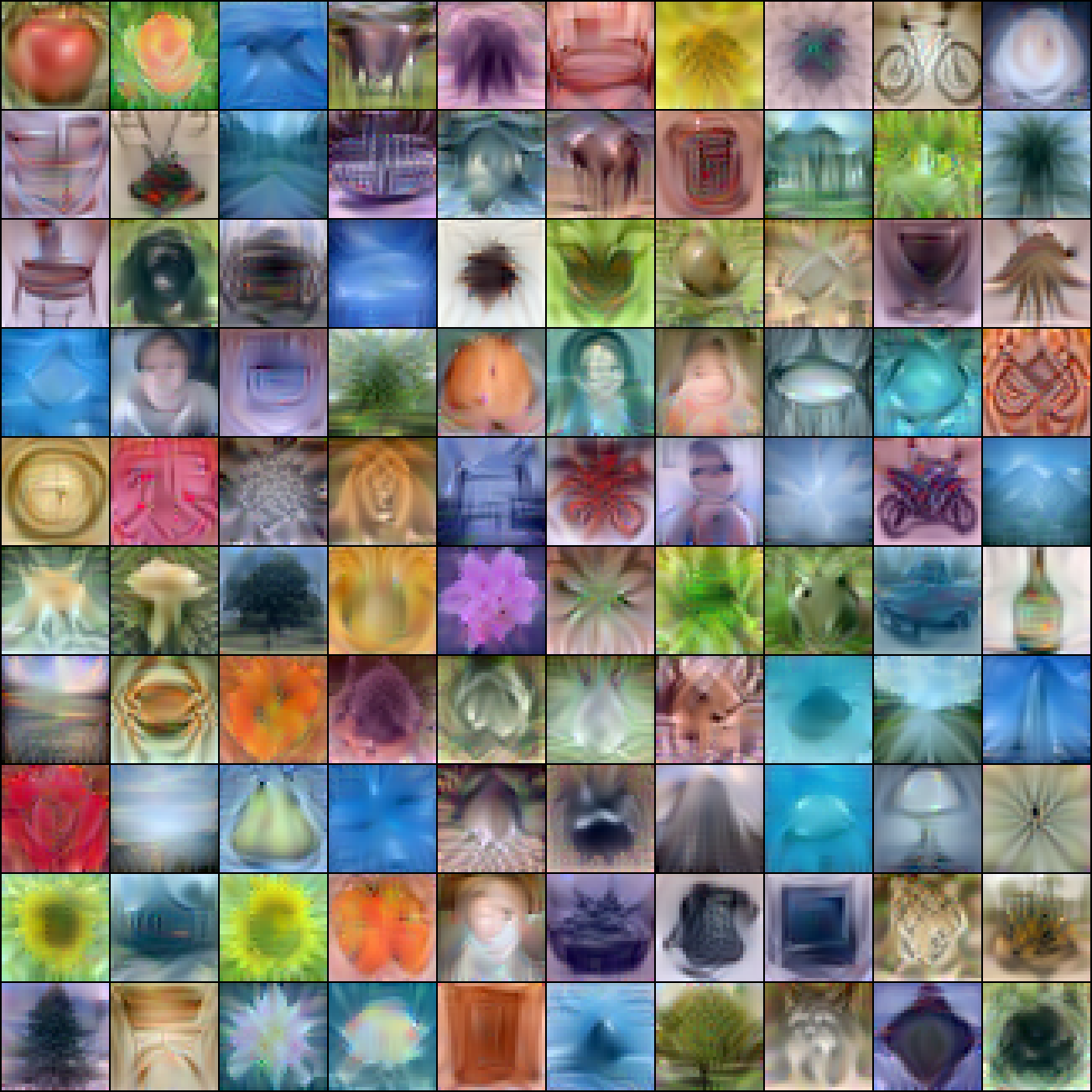}
    \caption{CIFAR100 IPC 1}
\end{figure*}

\begin{figure*}
    \centering
    \includegraphics[width=\textwidth]{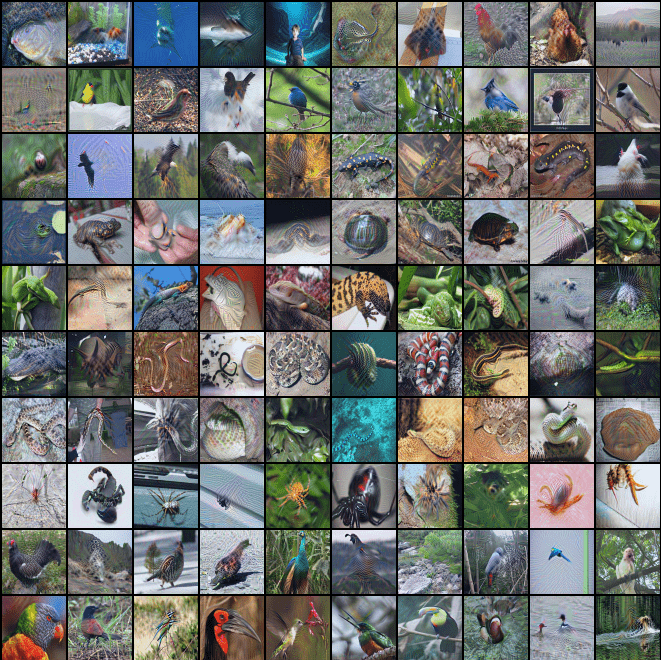}
    \caption{ImageNet-1K IPC 1, Class ID[0:99]}
\end{figure*}
\begin{figure*}
    \centering
    \includegraphics[width=\textwidth]{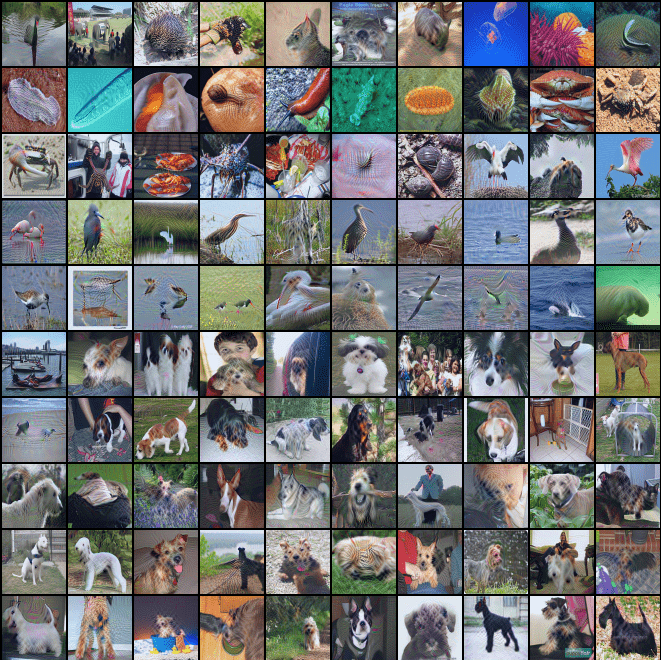}
    \caption{ImageNet-1K IPC 1, Class ID[100:199]}
\end{figure*}
\begin{figure*}
    \centering
    \includegraphics[width=\textwidth]{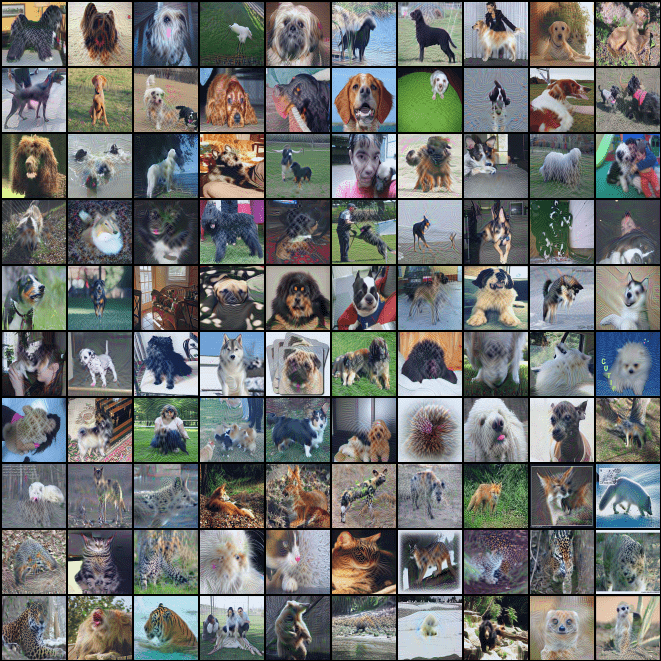}
    \caption{ImageNet-1K IPC 1, Class ID[200:299]}
\end{figure*}
\begin{figure*}
    \centering
    \includegraphics[width=\textwidth]{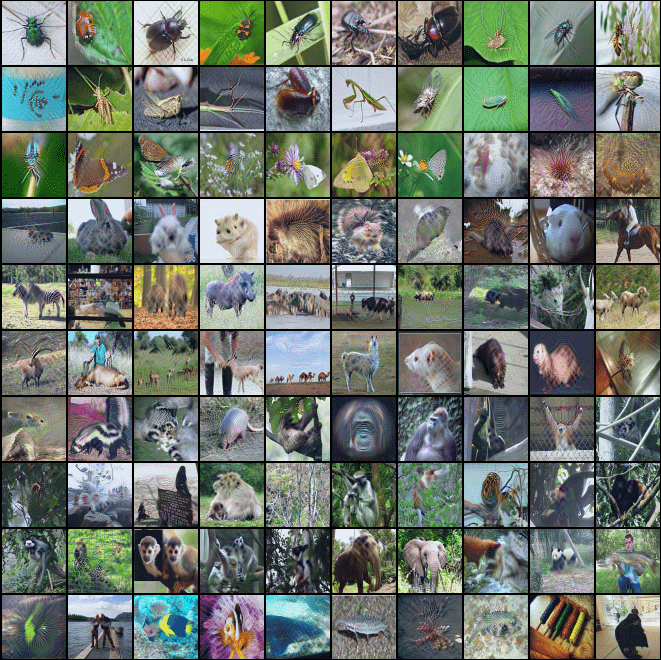}
    \caption{ImageNet-1K IPC 1, Class ID[300:399]}
\end{figure*}
\begin{figure*}
    \centering
    \includegraphics[width=\textwidth]{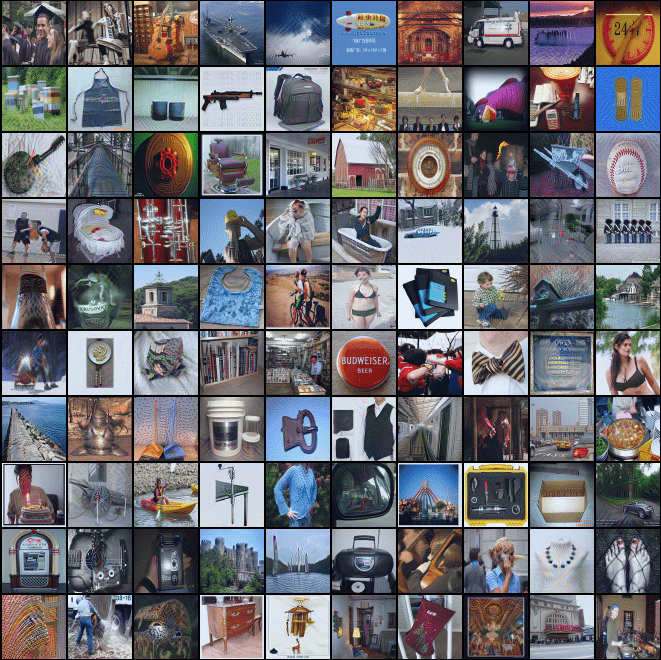}
    \caption{ImageNet-1K IPC 1, Class ID[400:499]}
\end{figure*}
\begin{figure*}
    \centering
    \includegraphics[width=\textwidth]{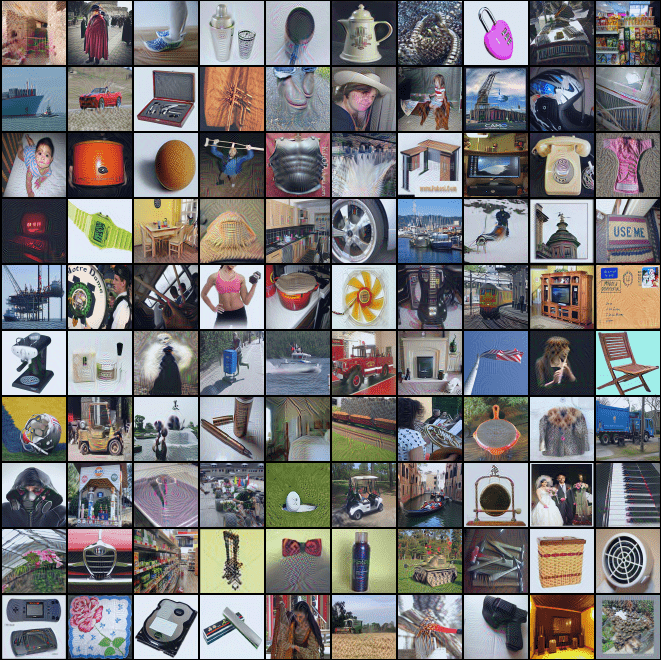}
    \caption{ImageNet-1K IPC 1, Class ID[500:599]}
\end{figure*}
\begin{figure*}
    \centering
    \includegraphics[width=\textwidth]{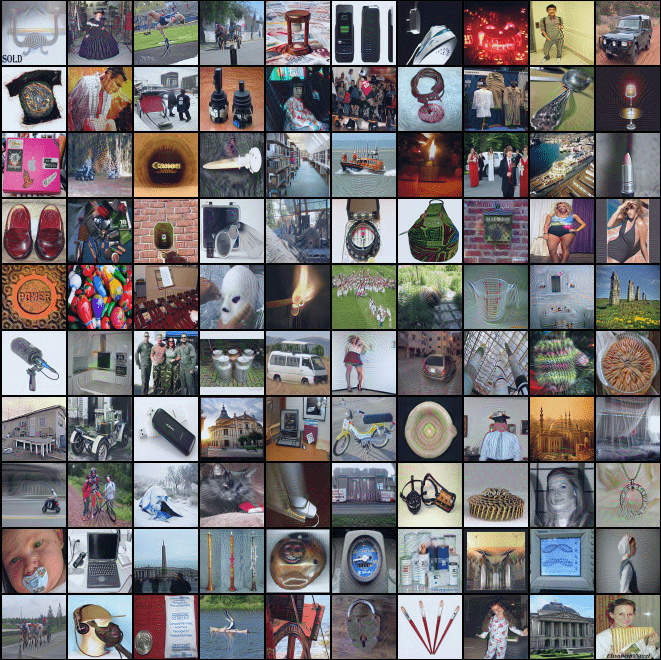}
    \caption{ImageNet-1K IPC 1, Class ID[600:699]}
\end{figure*}
\begin{figure*}
    \centering
    \includegraphics[width=\textwidth]{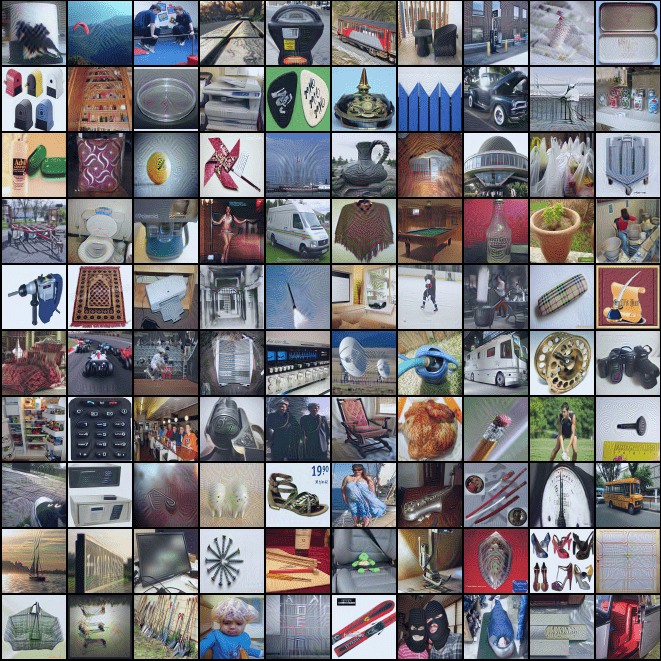}
    \caption{ImageNet-1K IPC 1, Class ID[700:799]}
\end{figure*}
\begin{figure*}
    \centering
    \includegraphics[width=\textwidth]{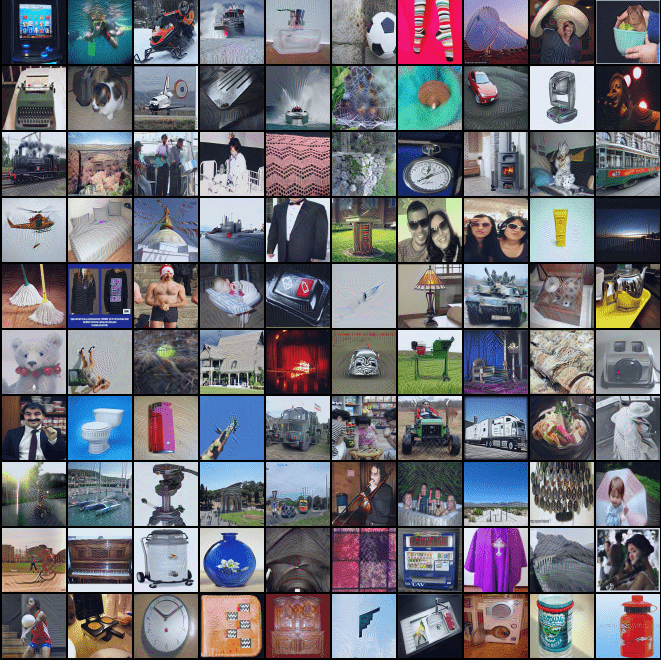}
    \caption{ImageNet-1K IPC 1, Class ID[800:899]}
\end{figure*}
\begin{figure*}
    \centering
    \includegraphics[width=\textwidth]{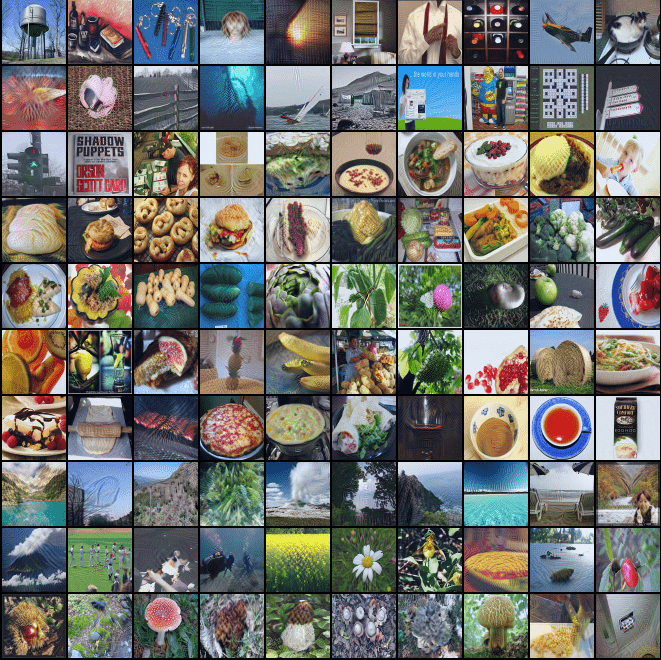}
    \caption{ImageNet-1K IPC 1, Class ID[900:999]}
\end{figure*}

\end{document}